# Internally-Balanced Magnetic Mechanisms Using Magnetic Spring for Producing Large Amplified Clamping Force

Tori Shimizu*, Kenjiro Tadakuma*,**,
Masahiro Watanabe, Eri Takane, Masashi Konyo, and Satoshi Tadokoro

*Abstract*— To detach a permanent magnet with a control force much smaller than its original attractive force, the Internally-Balanced Magnetic Unit (IB Magnet) was invented and has been applied to magnetic devices such as wall-climbing robots, ceil-dangling drones and modular swarm robots. In contrast to its drastic reduction rate on the control force, the IB Magnet has two major problems on its nonlinear spring which cancels out the internal force on the magnet: complicated design procedure and trade-off relationship between balancing precision and mechanism volume. This paper proposes a principle of a new balancing method for the IB Magnet which uses a like-pole pair of magnets as a magnetic spring, whose repulsive force ideally equals the attractive force of an unlike-pole pair exactly. To verify the proposed principle, the authors realized a prototype model of the IB Magnet using magnetic spring and verified through experiments its reduction rate is comparable to those of conventional IB Magnets. Moreover, the authors discussed and realized a robotic clamp as an application example containing proposed the proposed IB Magnets as its internal mechanism.

## I. INTRODUCTION

### A. Research Background

To maximize the operation time, permanent magnet is typically more effective to reduce electricity consumption than electromagnet when a mechanism utilizes a magnetic force to adhere to ferromagnetic surfaces and objects. However, since its attractive force is exerted permanently as its name says, permanent magnet requires a large control force to be detached from the target object, while this characteristic enables the magnet to keep attached to the target without any power consumption.

Typical ways for detaching a magnet, especially used for locomotion, are followings: applying a control force originally large enough or reduced enough to exceed the attractive force by a powerful actuator[1-3], inserting a separator between the magnet and the target object to decrease the attractive force gradually[4], canceling out the magnetic flux of the magnet by that of the electromagnet (which is called an electropermanent magnet)[5], disconnecting the magnetic circuit from the target object by switching the yoke[6]. Since these methods require an actuator with high torque, a gear box with high reduction ratio, or large current input for electromagnet, the advantage of the permanent magnet, ability to conserve energy for sustaining attraction, is ruined in these applications.

T. Shimizu, K. Tadakuma, M. Watanabe, E. Takane, M. Konyo, and S. Tadokoro are with the Graduation School of Information Sciences, Tohoku University, Sendai, Japan (phone: 022-795-7025; fax: 022-795-7023; e-mail: shimizu.tori@rm.is.tohoku.ac.jp, tadakuma@rm.is.tohoku.ac.jp, watanabe.masahiro@rm.is.tohoku.ac.jp, eri.takane@rm.is.tohoku.ac.jp, konyo@rm.is.tohoku.ac.jp, tadokoro@rm.is.tohoku.ac.jp, respectively).
* These authors contributed equally to this work.
** Corresponding authors

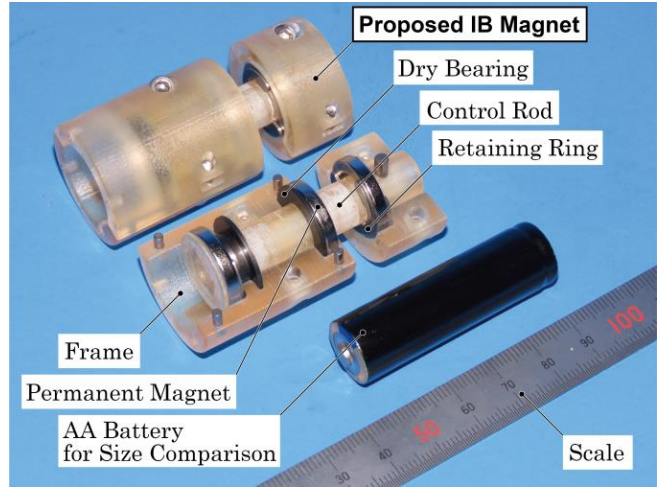

Figure 1. Proof-of-principle model of the proposed Internally-Balanced Magnetic Unit (IB Magnet) using magnetic spring.

### B. Internally-Balanced Magnetic Unit

To detach a permanent magnet with a control force much smaller than its original attractive force, the Internally-Balanced Magnetic Unit (IB Magnet for short) on Fig. 2 was invented[7] and has been applied to magnetic devices such as magnetic pad for climbing wall, anchor for ceil-dangling drones and connection unit of modular swarm robots[8-11].

The mechanism is composed of a permanent magnet for attraction held by the control rod and a nonlinear spring for internal balancing held by both the control rod and the mechanism frame. The spring is designed in a way it has a force-displacement repulsion characteristic $F_\mathrm{r}(x)$ identical but opposite in sign to force-displacement attraction characteristic $F_\mathrm{m}(x) = -F_\mathrm{r}(x)$ of the magnet so that the sum of these force-displacement characteristics, which is named the internal force $F_\mathrm{inter}(x) = F_\mathrm{r} + F_\mathrm{m}$, is calculated to be zero. Because the control rod the internal force is exerted on is then at the equilibrium point of force at any displacement $x$ from the attraction surface of the target object, ideally zero control force and control work are required for shifting the control rod to attach and detach the magnet, while the whole mechanism is still attracted to the target object by the counterforce exerted on the frame by the spring.

### C. Research Purpose

While its control force reduction is effective, the IB Magnet has not been applied widely because of problems on the nonlinear spring: its complicated design procedure and trade-off relationship between the precision of compensation and the mechanism volume. This paper proposes a principle of a new internal balancing method that solve these difficulties, realizes its prototype model to proof the proposed principle, and embodies a robotic clamp as its application example.

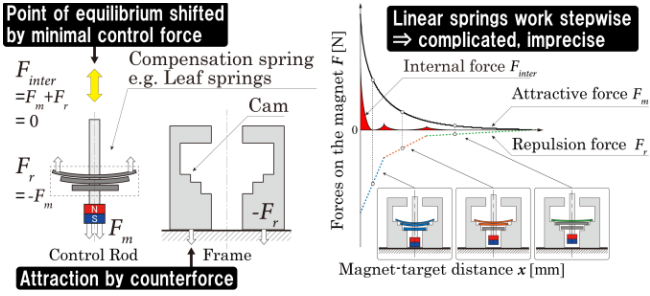
Figure 2. Principle diagram of a conventional Internally-Balanced Magnetic Unit (IB Magnet) and the characteristic of its nonlinear spring.

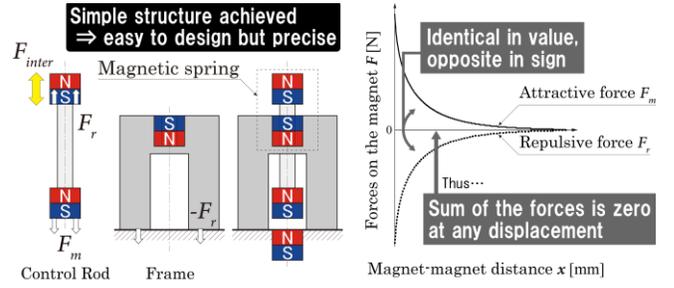
Figure 3. Principle diagram of the proposed Internally-Balanced Magnetic Unit (IB Magnet) and the characteristic of its magnetic spring.

## II. METHODS TO ACHIEVE INTERNAL FORCE COMPENSATION

### A. Conventional Method

The conventional nonlinear spring composed of multiple linear springs (leaf springs, for example) is shown in the Fig. 2 with its state transition according to the displacement of the control rod. To trace the force-displacement characteristic of the magnet that decreases inverse-proportionally to the square of the distance, the cam on the frame reduces the number of linear springs exerting force as the rod is pulled out further. The spring can be regarded as a mean of storing attraction work as elastic energy which gets released as repulsion work when control rod is to be pulled out to detach the magnet from the target object.

To optimize the nonlinear spring, an optimization should be conducted using the following equations. The loss of energy $\Delta E$ conversion from attraction work to elastic energy is calculated by attraction work can be calculated by Eq. (1), which integrates the out-of-balance force at every unit displacement $dx$ from the target surface $x = 0$ to the maximum stroke of the rod $x = x_{MAX}$. The optimization minimizes the $\Delta E$ by choosing spring constants $K_n (K_0 = 0)$ at tangent points $X_n (X_n < X_{n-1}, X_0 = 0)$ on the characteristic of the magnet and solving the Eq. (2) to find $x_n$, an intersection point of tangent lines with spring constants $K_n$ and $K_{n-1}$. The range $(0, x_n)$ defines the stroke of a single linear spring with spring constant $k_n$ calculated by using the Eq. (3) and (4).

$$\Delta E = \int_0^{x_{MAX}} \left\{ F_r(x) - \sum_n k_n x \ (x \le x_n) \right\} dx \quad (1)$$

$$x_n = \frac{(K_n X_n - K_{n-1} X_{n-1}) + \{F_r(X_n) - F_r(X_{n-1})\}}{K_n - K_{n-1}} \quad (2)$$

$$K_n = \left. \frac{dF_r(x)}{dx} \right|_{X_n} = \sum_n k_n \quad (3)$$

$$k_n = K_n - K_{n-1} \quad (4)$$

IB Magnet has enables drastic reduction rate on the control force, but its nonlinear spring has a serious problem of complicated design procedure. As shown above, the nonlinear spring composed of multiple linear springs require an optimization process which cannot be achieved by hand calculation, resulting in the difficulty in embedding in a practical mechanism. The design procedure of the spring thus becomes like following:

① Measure the attraction characteristic of the magnet.
② Define the stroke of the mechanism and the number of linear springs n to use. The more springs are used, the higher balancing precision (smaller $\Delta E$) will be achieved.
③ Draw a characteristic graph of the spring by placing n tangent lines to the characteristic of the magnet at $x_n$ and calculate their inclination equivalent to $K_n$.
④ Find out linear springs with spring constants close to $k_n$ and measure their precise free length.
⑤ Feedback the actual $k_n$ to the graph to define the starting positions $x_n$ of the springs and calculate the depth $X_n$ of the cam on the frame.

Furthermore, it is difficult to design a nonlinear spring which is both compact and precisely balancing because of the trade-off relationship between compensation precision and mechanism volume. While the more kinds of linear springs are needed to make the characteristic of the spring smoother and thus following the characteristic of the magnet with less deviation, it also makes the whole mechanism larger and heavier, becoming unsuitable for a mechanism after all, which reduces the control force a robotic wheel, a robotic arm or a drone with small actuators needs to exert.

### B. Proposed Method: Magnetic Spring

The authors invented a new internal balancing method of magnetic spring[12-13] as shown in the Fig. 3. It utilizes the property of a pair of magnets: the repulsive force characteristic of a like-pole pair, which is called a magnetic spring, is ideally identical but opposite in sign to the attractive force of an unlike-pole pair. Just by aligning two pair of magnets in a same distance the sum of their force always equals zero for any displacement on the stroke.

The proposed magnetic spring has apparently no more complicated design procedure than the conventional spring:

① Measure the attraction characteristic of a pair of the magnets.
② Define the stroke of the mechanism which is also equal to the maximum displacement of the springs for the magnetic spring.

The proposed IB Magnet using magnetic spring is so innovative that it can be composed just by identical magnets but drastically simplifies the design of the basic structure while sustaining or even improving the compensation precision. The magnetic spring gives advantage of endless rotatability of the control rod relative to the frame because of the point symmetry and contactlessness.

## III. EMBODIMENT OF THE PROTOTYPE MODEL

### A. Conceptual Design

Based on the proposed principle, the authors designed a proof-of principle prototype model of the IB Magnet using magnetic spring as shown in the Table I and Fig. 4. Ring magnets magnetized in the thickness direction are selected under the condition that a tube-like rod and dry bearings can be placed through the holes for lubrication and future expandability such as controlling the rod by the side of the attraction magnet, rotating the control rod relative to the frame and/or target surface, and ducting fluids. Their surface to the corresponding surface (its pair magnet or the target object) is shielded by a cover with thickness of 1 [mm] to avoid the ferromagnetic sand. The stroke is defined to be 7.5 [mm], where a material testing machine (Instron, 3343) measured the attractive force between two magnets to be small enough 0.5 [N] under the covered condition same as the prototype model.

### B. Building the Prototype Model

The Fig. 1 shows the embodied proof-of-principle prototype model of the proposed IB Magnet using magnetic spring. While it is stiff enough for the proof of principle, a 3D-printed acrylic material (Keyence, AR-M2) is used for the entire structure including the rod for the sake of rapid prototyping for fluent proof of principle in this report and may well affect the stretch strength, compression strength and bearing precision of the mechanism. Materials with higher stiffness such as polycarbonate, CFRP nylon, and aluminum in the future research.

## IV. FUNDAMENTAL EXPERIMENT

To proof and evaluate the internal force compensation effect of the proposed IB Magnet using magnetic spring, the authors conducted an extension examination using the embodied prototype model and material testing machine as shown in the Fig. 5. The IB Magnet at the initial condition is placed on and attached to the target acrylic plate with the target magnet buried under 1 [mm] depth, and either its frame or control rod is pulled by the pulling jig and isolated T-shape hook connected to the examination machine at the rate of 0.5 [mm/sec] and the displacement of 0 to 15 [mm] (twice the stroke) to measure the control force required to detach the IB Magnet pulling the corresponding part of the mechanism.

The Fig. 6 and 7 shows the 5-time average result of the measurements. (a)When the ascending T-shape hook touches the jig connected to the frame, the control force suddenly reaches the highest net value of 8.4 [N], identical to the maximum attractive force of the original magnet, and then the detach of the magnets begins the gradual decrease of the control force to the weight of the whole mechanism and the jig. (b)When the jig is connected to the control rod, the control force likely increases but the highest net value 1.1 [N] is apparently reduced to 13.0 [%] of the value at frame pulling. The control force then decreases to the total weight with one rising edge when the rod touches the frame at its maximum stroke. The control force reduction ratios of conventional springs are reported 11.8[%] with 6 kinds of coil springs and 15.4[%] with a ring-shaped Neidhart rubber spring[14], so the experiment successfully validated the effectiveness of the proposed magnetic spring and likely reduction ration with much simpler structure and easier design procedure.

TABLE I. SPECIFICATION OF THE PROTOTYPE MODEL OF THE PROPOSED IB MAGNET USING MAGNETIC SPRING

| | Type Number | HXCW18-12-3 |
|---|---|---|
| Magnet | Outer Diameter | 18[mm] |
| | Inner Diameter | 12[mm] |
| | Thickness | 3[mm] |
| | Weight | 2.9[g] |
| Diameter | | 28.0[mm] |
| Length | | 58[mm] |
| Stroke | | 7.5[mm] |
| Weight | | 44.3[g] |

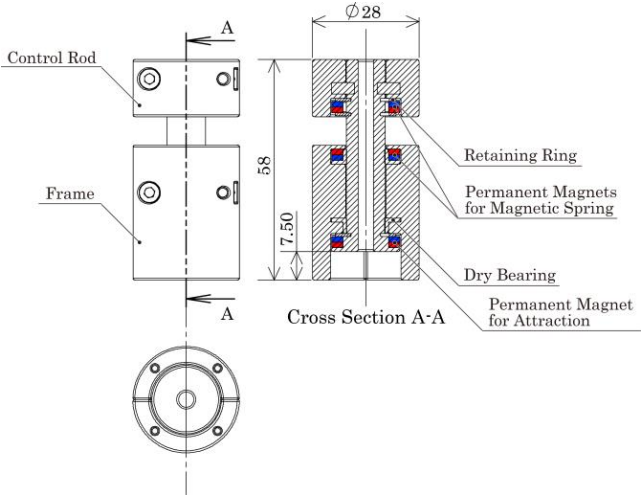

Figure 4. Design sketch of the proposed IB Magnet using magnetic spring

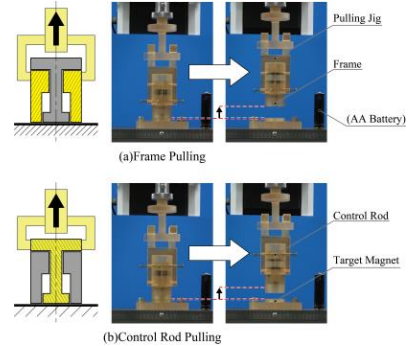

Figure 5. Setting of the experiment for evaluating the internal force compensation effect of the proposed IB Magnet using magnetic spring

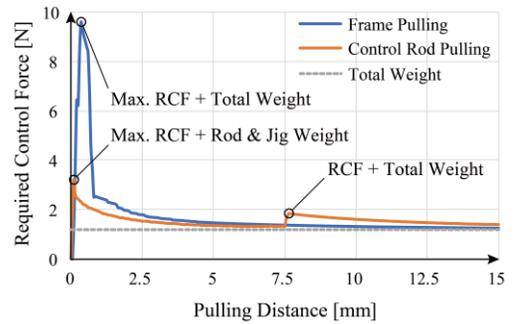

Figure 6. Result of the experiment for evaluating the internal force compensation effect of the proposed IB Magnet using magnetic spring, distance-force characteristics

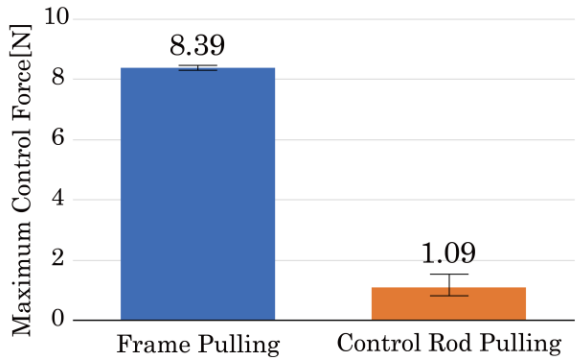

Figure 7. Result of the experiment for evaluating the internal force compensation effect of the proposed IB Magnet using magnetic spring, direct comparison of maximum net control forces (weight subtracted)

## V. APPLICATION OF THE PROPOSED IB MAGNET TO A ROBOTIC CLAMPING MECHANISM

### A. Application Concept

The authors have proposed an idea of "equilibrium-floating force-displacement converter", a mechanism which is composed of a spring with spring characteristic $F_s(x)$ and an "inverse-spring" with spring characteristic $F_{AS}(x) = -F_s(x)$ to balance the internal force $F_{INTER}(x) = F_s(x) + F_{AS}(x)$ exerted on the connection point of these springs. It is thus capable of adjusting the force exerted on each end of the springs by shifting the equilibrium point with a minimal control force[15]. In other words, the force of a small, lightweight and minimal actuator can be amplified to a much larger force just by inputting a parallel displacement.

The IB Magnet can be counted among the force-displacement converters by regarding the attraction magnet and control rod as the inverse-spring and equilibrium point of the force, respectively. $F_s(x)$ is now $F_r(x)$ and $F_{AS}(x)$ becomes $F_m(x)$ in this view. The end of the inverse-spring corresponds to the attractive force exerted on the frame, which is equal to the weight the ferromagnetic object the IB Magnet can lift up at that point on the stroke of the rod.

The authors thus decided to embed the IB Magnet, especially the one using the proposed magnetic spring to make the design simpler, in a mechanism that requires control of a large force by including the attraction target inside the mechanism itself.

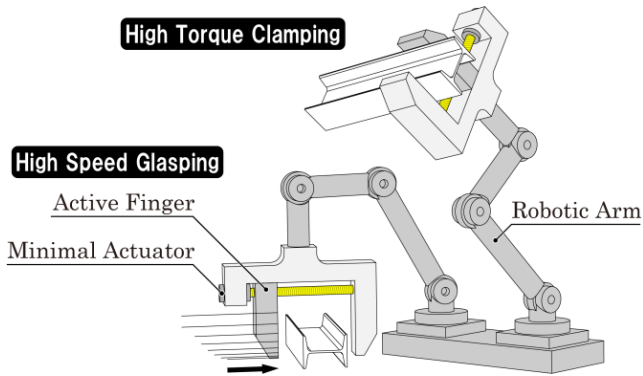

Figure 8. Conceptual diagram of an ideal claming mechanism for a robotic arm with state transition function

### B. Clamping Mechanism using IB Magnet

Considering the response time, operation time and power consumption, ideal robotic grippers (or actuator-driven clamps or vises, essentially) widely used such as in industry and disaster rescue should be able to change the reduction states just by one actuator as shown in the Fig. 8: high-speed state until the finger touches the object to be able to complete a grasping quickly and the high-torque state after the finger touches the object to sustain the grasping position firmly. Conventionally, this state transition has been achieved by using sensors that tells the existence of the grasping object between the fingers and regulating the current for the actuator that drives an active finger, but this electronical method requires a certain processing load on the robotic system. To relief the load on control there have been proposed passive reduction mechanisms that use a toggle or a clutch, mechanical joints with dead centers that restrict the further movement of the actuator position, to detect the load on the finger when touch to the object occurs and switch the reduction state of the actuator[16-18]. However, they are not capable of adjusting the clamping force because these switching mechanisms allows only binary state of clamping: either at 0 or maximum force. This is troublesome when the robotic system handles fragile objects, .

To solve these problems, the authors newly propose a clamping mechanism that applies the IB Magnet to grasp a target object by the attraction movement between itself as an active finger and the ferromagnetic surface as a fixed finger. Since the magnet enlarges the attractive force spontaneously and gradually as it gets closer to the ferromagnetic surface, the mechanism can adjust the clamping force simply by shifting the control rod of the IB Magnet.

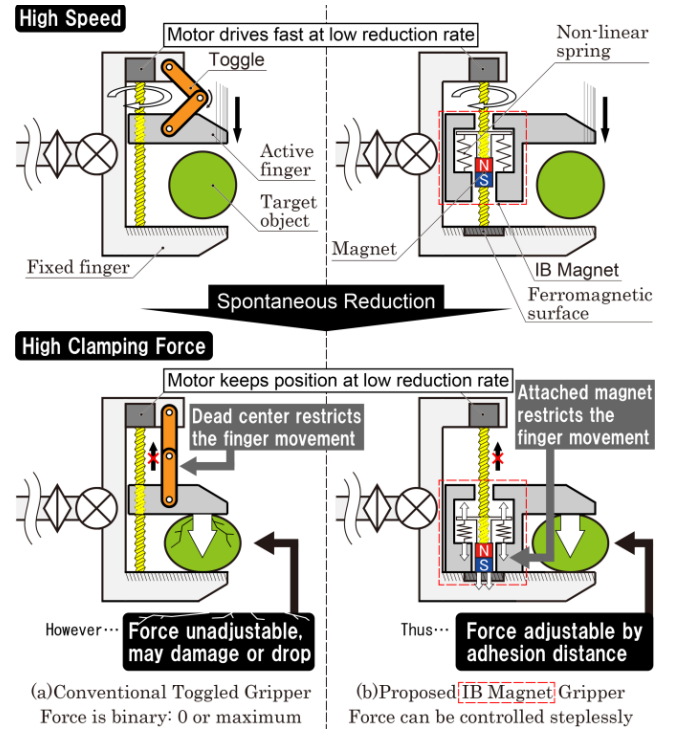

Figure 9. Problems of a conventnional mechanical clamp and the proposed clamp mechanism using IB Magne

## C. Enlargement of the IB Magnet using magnetic spring

Since the former model of the Fig. 1 uses magnets with small attractive force, the authors created a larger IB Magnet using magnetic spring as shown in the Fig. 10 and the Table 2 to embed in the mechanical clamp. The structure and the control rod are composed of 3D-printed ABS material (Stratasys, ABS-P430) and non-magnetic stainless steel (SUS304), respectively. The result of an experiment, conducted under the same condition as the chapter IV except that the rate of stretch is 1.0 [mm/sec] since the stroke became longer than the former model, shows that the principle of the IB Magnet using magnetic spring is valid at a larger scale but with a lower compensation precision (to be discussed later).

The control rod of the former model used to hold the magnets by sandwiching one between retaining rings or between a retaining rings and a flange of the rod. Since this design is effective for avoiding magentic sand but drastically reduces the maximum attractive force, this larger version acquired jigs tied by screws to the rod which hold the magnets by press fitting aided with high shear strength glue to achieve zero-distance adhesion. The target object at the experiment and the clamp is set to be an aluminum plate with a magnet embedded on its surface, identical to the magnets used.

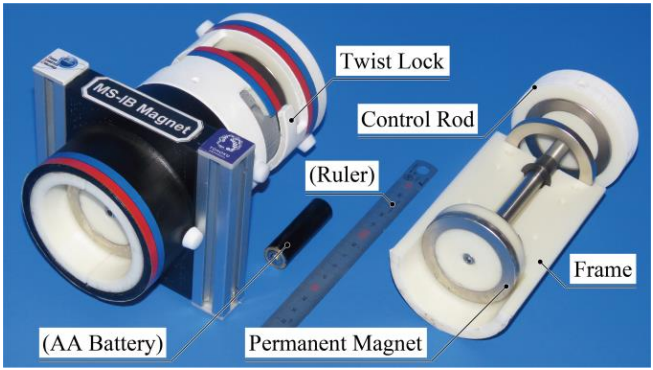

Figure 10. Enlarged version of the IB Magnet using magnetic spring

TABLE II. SPECIFICATION OF THE ENLARGED VERSION OF THE IB MAGNET USING MAGNETIC SPRING

| | Type Number | NOR391 |
|---|---|---|
| Magnet | Outer Diameter | 54 [mm] |
| | Inner Diameter | 38 [mm] |
| | Thickness | 5 [mm] |
| | Weight | 40.8 [g] |
| Diameter | | 80 [mm] |
| Maximum Length | | 160 [mm] |
| Stroke | | 20 [mm] |
| Weight | | 683.4 [g] |

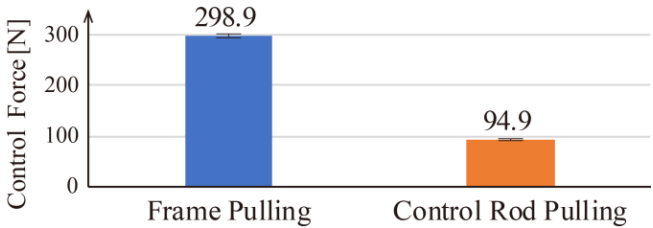

Figure 11. Result of the experiment for evaluating the internal force compensation effect of the enlarged version of the IB Magnet using magnetic unit, direct comparison of maximum control forces

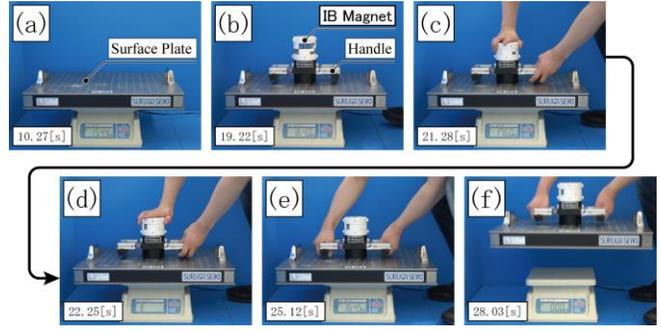

Figure 12. Demonstration of how to use the enlarged version of the IB Magnet using magnetic spring, a surface plate of 15.4 [kg] as a target object

In addition, to avoid the sudden and unintended release phenomenon of the control rod resulting in the forced separation of the mechanism and the target object, which occasionally happens because the internal force is too balanced to be easily shifted by small shake or vibration, a lock mechanism which can be locked and unlocked by twisting was installed in this version as used in the Fig. 12. This ensures not only the complete adhesion state but also the complete detached state of the magnet to/from the target object. A bearing is inserted between the control rod and the repulsion magnet to make the attraction magnet independent of the lock's rotation. The authors, other than that, got a reserve thought on this disadvantageous phenomenon and proposed a jumping mechanism using the inertia produced by the powerfully released rod, as reported in authors' other reports on the applied researches of the IB Magnet [19-20].

## VI. EMBODIMENT AND EXPERIMENT OF THE PROPOSED IB MAGNET CLAMPING MECHANISM

### A. Conceptual Design

Based on the proposed principle, the authors designed a hand-powered proof-of-principle prototype model of the proposed IB Magnet clamping mechanism as shown in the Fig. 13. The grasping width is fixed to 35 [mm], the height of a loadcell (Kyowa Sangyo, LUR-A-2KNSA) used in the later experiment. The mechanism contains the IB Magnet of the Fig. 10 and other structural parts are composed of aluminum frame. The IB Magnet can be shifted upward for more than its stroke to fit to objects thicker than the designed grasping width.

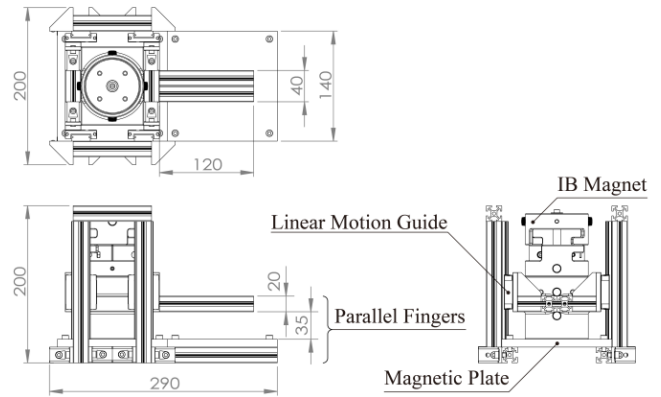

Figure 13. Design sketch of the proposed IB Magnet clamping mechanism embedding the enlarged version of the IB Magnet using magnetic spring

## B. Building the Prototype Model

The Fig. 14 shows the embodied proof-of-principle prototype model of the proposed IB Magnet clamping mechanism.

## C. Clamping Experiment

To proof and evaluate the spontaneous reduction effect by the internal force compensation of the proposed IB Magnet clamping mechanism using magnetic spring, the authors conducted a clamping examination using the embodied prototype model as shown in the Fig. 15 and the result in the Fig. 16. The clamping force is originated from the strain of the mechanism and the target object, so the fixed finger of the bottom is substituted by a plate 4 [mm] thinner than the magnetic plate, the adhesion target of the IB Magnet, and raised 5 [mm] by inserting an aluminum plate to make the situation that the mechanism grasps an object 1 [mm] thicker than its minimum inter-finger distance so that it can exert force properly.

The experiment advanced as following: (b)The loadcell was inserted between the parallel fingers, (c)and the initial load on the loadcell exerted by the weight of the active finger was measured to be 12.9 [N], the gravity bias of the clamping force. (d)A weight of 9.68 [kgf], the maximum control force 94.9 [N] required to shift up and down the control rod as shown in the Fig. 11, is applied on the rod in a fully pulled-out state and the net clamping force without the spontaneous reduction was measured to be 18.5 [N]. (e)The control rod was shifted down by hand, (f)and now the net clamping force with the spontaneous reduction by the magnet was measured to be 36.9 [N], 2.0 times the original clamping force at (d), showing that The experiment successfully validated the effectiveness of the proposed IB Magnet clamping mechanism.

## VII. DISCUSSION

Both the proof-of-principle model of the proposed IB Magnet using magnetic spring and its enlarged version still had measurable compensation impreciseness resulting in non-zero control force when pulling out the control rod, and a larger deviation of the balance was observed at the enlarged version. The major possible cause of this phenomenon would be the difference between the shape of the magnetic circuits, and thus the exerted magnetic force, of a like-pole pair of magnets and a unlike-pole pair of magnets at zero-distance alignment. The deviation of the attraction and repulsion characteristics around the origin may have become larger and more unignorable as the volume and the magnetic flux got larger. The authors are going to conduct a static magnetic field analysis to comprehend this nature and introduce additional internal force adjusting mechanisms such as one that gives an offset displacement or constant force on the attraction magnet so that the internal force at the origin can be set to zero and at the other displacement acceptably positive or negative which is in the range of the force the minimal actuator can hold.

Another problematic feature of the proposed IB Magnet using magnetic spring is that the magnetic spring always produces a larger repulsive force than the attraction magnet does a attractive force, because the attraction force between the attraction magnet and a target object cannot be the largest as long as the target is the identical magnet to the magnets used in the mechanism. This problem does not occur in conventional IB Magnets since their springs can be designed for each known target object individually, but the merit of magnetic spring is worth emdbedding in the applications where they can contain a magnet as attraction target inside. Likewise, while the proof-of-principle model of the proposed IB Magnet clamping mechanism is just a prototype and cannot grasp thinner or much thicker objects than its designed grasping width, it still can be directly applied to mechanisms with known and fixed target objects and take advantage of the spontaneous redction effect: e.g. a brake for wheels.

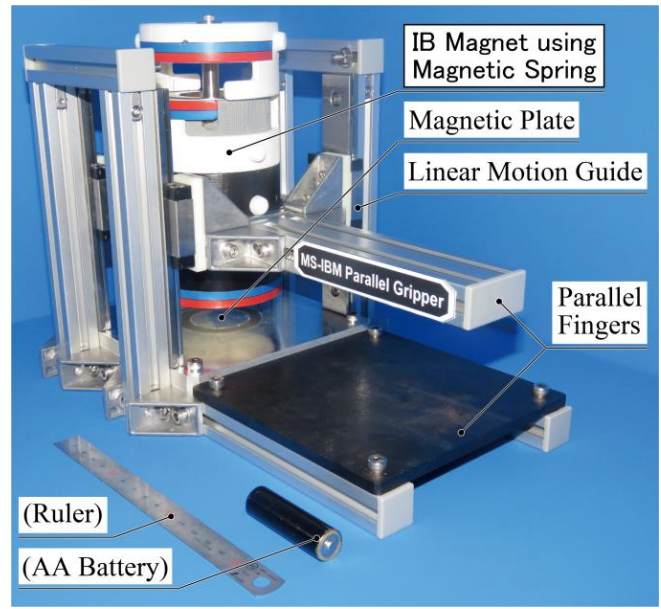

Figure 14. Proof-of-principle model of the proposed IB Magnet clamping mechanism using magnetic spring

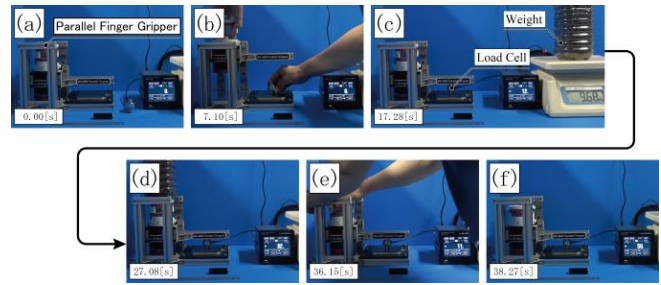

Figure 15. Grasping experiment for evaluating the spontaneous reduction effect of the proposed IB Magnet clamping mechanism using magnetic spring

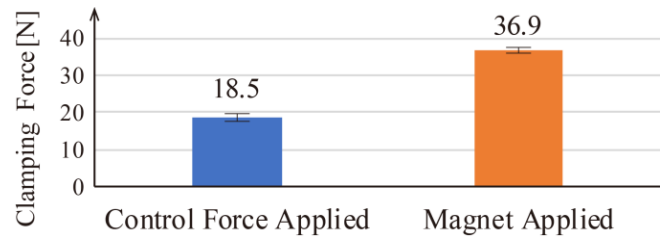

Figure 16. Result of the experiment for evaluating the spontaneous reduction effect of the proposed IB Magnet clamping mechanism using magnetic spring

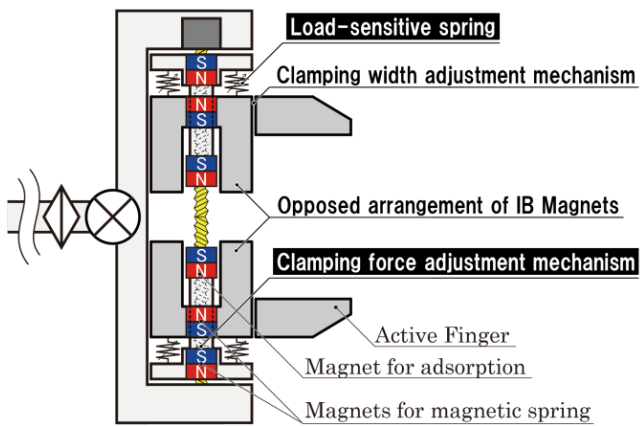

Figure 17. Conceptual design of the future versioin of the proposed IB Magnet clamping mechanism using magnetic spring

VIII. CONCLUSION AND FUTURE PROSPECTS

In this paper, the authors proposed a principle of a new balancing method using a magnetic spring to solve the problems of the conventional nonlinear spring used for the IB Magnet: complicated design procedure and trade-off relationship between the mechanism volume and the force compensation precision. To verify the proposed principle, the authors realized a prototype model of the IB Magnet using magnetic spring and conducted a experiment to show that its reduction rate is comparable to those of conventional IB Magnets while its structure became much simpler. The authors enlarged the IB Magnet using magnetic spring to embody a robotic clamp embedding it as a equilibrium-floating force-displacement converter and also verified its effectiveness on spontaneous force reduction effect as a clamping force amplifier.

In the future reports, the authors are planning to implement the proposed IB Magnet clamping mechanism in a way shown in the Fig. 17. By oppositely arranging a pair of the IB Magnets with opposite magnetized direction, both fingers will be active and the deviation of the target object from the center of the grasping point will be allowed with more acceptance. The control rod will have a clamping force adjustment mechanism to be held the shaft at any displacement to keep a clamping force at that point. In addition, a load-sensitive spring will be needed to compensate the weight component on the internal force that differs by the orientation. Furthermore, clamping width should be able to be adjusted by a mechanism between the finger and the frame but with a high stiffness and may well be included in the load-sensitive spring so that the active degree of freedom still remains to one: the actuator that drives the screw shaft connected to the control rods of the IB Magnets.

ACKNOWLEDGEMENT

This work was supported by the "Impulsing Paradigm Change through Disruptive Technologies Program (ImPACT)", a research division designed by the Council for Science, Technology and Innovation of Japan, and was consigned via Japan Science and Technology Agency. The authors are grateful to their kind financial support.